\documentclass[10pt,a4paper]{article}

\usepackage[T1]{fontenc}
\usepackage[utf8]{inputenc}
\usepackage{authblk} 
\usepackage[margin=1in]{geometry}
\usepackage{graphicx}
\usepackage{booktabs} 
\usepackage{amsmath,amssymb}
\usepackage{url}
\usepackage{natbib}
\usepackage{xcolor}

\usepackage{microtype}
\usepackage{placeins} 
\usepackage{titlesec} 
\usepackage{multirow}
\usepackage{float}

\titleformat{\section}
  {\normalfont\Large\bfseries\raggedright}{\thesection.}{1em}{}
\titleformat{\subsection}
  {\normalfont\large\bfseries\raggedright}{\thesubsection.}{1em}{} 
\titleformat{\subsubsection}
  {\normalfont\normalsize\bfseries\raggedright}{\thesubsubsection.}{1em}{} 
\usepackage{hyperref}
\hypersetup{
    colorlinks=true,
    linkcolor=blue,
    filecolor=magenta,      
    urlcolor=blue,
    citecolor=blue,
    pdftitle={Growing Transformers: Modular Composition and Layer-wise Expansion on a Frozen Substrate},
    pdfauthor={Andrey Bochkov},
}

\title{Growing Transformers: Modular Composition and Layer-wise Expansion on a Frozen Substrate}

\author[1]{Andrey Bochkov\thanks{Corresponding author: andrey.bochkov@gmail.com}}
\affil[1]{Moscow Institute of Physics and Technology (MIPT), Moscow, Russia}

\date{} 

\begin{document}

\maketitle

\begin{abstract}
We study a constrained training regime for decoder-only Transformers in which the token interface is fixed, previously trained dense blocks are not reopened, and the active trainable parameter set is kept approximately constant as depth grows. Starting from a shallow model, we stack new blocks and train only the newest blocks and the LM head; optional LoRA phases provide limited global readjustment under the same active-parameter budget. The paper asks a feasibility/tradeoff question, not whether this regime matches tuned monolithic pretraining.

In a common-protocol 9-layer study on a frozen Unicode substrate, the constructive frozen-Unicode model uses 105.0M active trainable parameters, compared with 180.5M for the interface-matched monolithic frozen baseline and 247.6M for the fully trainable monolithic baseline. We then consider an extreme fixed interface: each token is represented only by a frozen 16-dim binary token-ID code, deterministically lifted to $d_{\text{model}}$, so the resulting token embedding matrix in $\mathbb{R}^{d_{\text{model}}}$ has rank at most 16. Even in this setting, continued growth remains viable. In a 68.9B-token run on FineWeb-Edu + Cosmopedia, a 16-layer 269.7M model trained above this fixed interface reaches 28.92\% MMLU after an interleaved LoRA stage. Reported final metrics are measured after merging the last-stage LoRA adapters into the 269.7M base model. Because the data mixture changes across stages in this long-horizon run, we interpret it as a viability demonstration rather than a clean causal comparison.

Overall, the evidence supports a narrow claim: useful continued learning can proceed above a frozen minimal interface under a bounded active trainable-parameter budget, with a clear tradeoff against dense monolithic training in final perplexity.
\end{abstract}

\section{Introduction}

Large-scale language model training is often constrained by optimizer-state memory rather than by parameter count alone. In standard dense monolithic pretraining, optimizer states must be maintained for all trainable parameters, which can make scaling difficult under a fixed accelerator memory budget. This motivates a simple question: how much continued learning remains possible if only a bounded subset of the model is trainable at any given time?

This paper follows \citet{bochkov2025emergent}, which showed that useful language modeling can arise even when the input embedding layer is fully frozen and non-semantic. Here we ask a narrower follow-up question. Once the token interface is fixed, can model depth continue to grow without reopening previously trained dense blocks, while keeping the active trainable parameter set approximately constant? We study this as a feasibility and tradeoff question rather than as a claim of superiority over tuned dense monolithic pretraining.

Our training regime is deliberately restrictive. We begin with a shallow decoder-only Transformer, train it, freeze the trained dense blocks, stack new blocks on top, and continue training only the newest blocks together with the LM head. For some longer runs we interleave LoRA stages across all layers, but only under the same active trainable-parameter budget; no stage performs dense full-model reopening. The paper therefore studies a bounded-active-budget regime rather than generic progressive growth.

A central stress test in this work is an extreme fixed token interface. In the $n_{\text{embed}}=16$ setting, each token is represented only by a frozen 16-dim binary code corresponding to its token ID, followed by a deterministic lift to $d_{\text{model}}$. This setup is intentionally minimal: the induced lookup table in $\mathbb{R}^{d_{\text{model}}}$ has matrix rank at most 16, so it is not a learned high-dimensional embedding in disguise. If continued learning remains possible above such an interface, then semantically rich trainable embeddings are not a prerequisite for useful depth growth.

The paper reports two main empirical views of this regime. First, we run a controlled 9-layer common-protocol comparison on a frozen Unicode substrate and quantify the quality--memory tradeoff against monolithic baselines. Second, we run a longer-horizon 16-layer experiment on FineWeb-Edu + Cosmopedia above the 16-dim binary fixed token-ID interface, using interleaved LoRA stages under the same active trainable-parameter budget. 

The main conclusion is narrow. Dense monolithic training remains the stronger quality baseline when its systems cost is acceptable. However, useful continued learning can still proceed above a fixed minimal interface while keeping the active trainable set bounded, and this remains true even in the extreme 16-dim binary token-ID setting. Model, tokenizer files, and usage examples are available at \url{https://huggingface.co/Bochkov}.

\subsection{Contributions}

We make the following limited claims:

\begin{itemize}
    \item We define and study a constrained training regime that combines (i) a fixed deterministic token interface, (ii) freezing of previously trained dense blocks, and (iii) an approximately fixed active trainable-parameter budget during depth growth.

    \item In a common-protocol 9-layer comparison, active trainable parameters drop from 180.5M to 105.0M relative to the interface-matched monolithic frozen baseline (and from 247.6M to 105.0M relative to the fully trainable monolithic baseline), while final training perplexity rises from 3.71 to 4.12 relative to the interface-matched monolithic frozen baseline.

    \item We show that continued growth remains viable above a minimal 16-dim binary frozen token-ID code, where the effective input subspace before the first trainable block has dimension at most 16.

    \item We report a long-horizon 68.9B-token run with interleaved LoRA global readjustment under the same active trainable-parameter budget, and we make the LoRA target modules and ranks explicit for reproducibility.
\end{itemize}

\section{Related Work}

\subsection{Layer-wise training and model growth}

Greedy layer-wise training has a long history in deep learning, including early work on deep belief nets and deep autoencoders \citep{hinton2006fast,bengio2007greedy}. Later work studied explicit model growth and function-preserving transformations, notably Net2Net \citep{chen2016net2net}, where a larger model is initialized from a smaller one in a way that preserves or approximately preserves the original function.

For Transformers, progressively growing depth has been explored in several
settings, including \emph{Efficient Training of BERT by Progressively Stacking}
\citep{gong2019progressive}, \emph{Progressively Stacking 2.0}
\citep{yang2020progressive}, \emph{Learning to Grow Pretrained Models for
Efficient Transformer Training} (LiGO) \citep{wang2023ligo},
\emph{Stacking Your Transformers} \citep{du2024stacking},
\emph{LLaMA Pro} \citep{wu2024llamapro},
\emph{Reuse, Don't Retrain} \citep{parmar2024reusedontretrainrecipe},
 \emph{Curriculum-Guided Layer Scaling} \citep{singh2026curriculumguidedlayerscalinglanguage} and \emph{LESA: Learnable LLM Layer Scaling-Up}\citep{yan2025lesa}.
These papers already show that model depth can often be expanded during
training more cheaply than training the final architecture from scratch, and
some variants also freeze previously trained layers during part of the
schedule. We therefore do not claim progressive stacking itself as a novel idea.

\subsection{Positioning of this work}

The present paper studies a narrower regime than generic growth or stacking. Our focus is on the combination of four constraints: (i) the token interface is fixed and deterministic throughout training, (ii) previously trained dense blocks remain frozen, (iii) the active trainable-parameter set is kept approximately constant as depth grows, and (iv) when used, LoRA serves only as a budget-preserving global readjustment stage rather than dense full-model reopening. The contribution is therefore an empirical study of this constrained regime, not a claim to outperform copy-based or function-preserving growth strategies.

\subsection{Memory-efficient training and parameter-efficient adaptation}

Large-model training is often limited by optimizer-state memory, motivating sharded optimizers, offloading, and lower-precision optimizer states \citep{rajbhandari2020zero,dettmers2022eightbit}. Our approach is complementary. Instead of storing optimizer state for all dense parameters, we restrict which parameters are trainable at each stage. Adapter methods and LoRA similarly reduce the trainable set while keeping a large base model frozen \citep{pfeiffer2021adapterfusion,hu2022lora}. In our setting, however, LoRA is not used as a standard downstream fine-tuning recipe; it is used during continued pretraining as a budget-preserving global adjustment stage between growth phases.

\subsection{Frozen and non-semantic token interfaces}

This work directly builds on \citet{bochkov2025emergent}, which showed that useful language modeling can arise with a fully frozen visual Unicode embedding layer. The current paper asks a follow-up question: once the token interface is fixed, can depth continue to grow without reopening the previously trained dense stack? We study this both on the visual Unicode substrate and on an even stronger stress test, a frozen 16-dim binary token-ID code. In the latter case, the induced lookup table in $\mathbb{R}^{d_{\text{model}}}$ has matrix rank at most 16, so the setup is not a learned high-dimensional embedding in disguise.

\section{Method}

\subsection{Problem setting}

We study decoder-only Transformers under an explicit constraint on the \emph{active} trainable parameter set. Let $B_{\text{act}}$ denote the target budget of trainable parameters at a given stage. The goal is to increase model depth while keeping the optimizer-state footprint approximately bounded by ensuring that only a subset of the model is trainable at any time.

At each stage, the full model participates in the forward pass, but only the newest blocks and the LM head are updated. Previously trained dense blocks remain frozen. In some runs we additionally interleave LoRA stages that reallocate the same active-parameter budget to adapters across all layers, again without dense full-model reopening.

\subsection{Fixed token interfaces}

We consider two fixed token interfaces. The first is the frozen visual Unicode interface introduced in \citet{bochkov2025emergent}. The second is a stronger stress test: a frozen 16-dim binary token-ID code. In both cases, the interface is fixed throughout training and is not adapted jointly with later layers.

The purpose of these fixed interfaces is not to claim that they are universally preferable to learned embeddings. Rather, they let us ask a sharper question: once the bottom interface is fixed, how much continued learning can still be achieved by training only newly added upper blocks under a bounded active trainable budget?

\subsection{Clarifying the 16-dim binary token-ID interface}

In the $n_{\text{embed}}=16$ setting, each token $t$ is assigned a fixed binary code
\[
b(t)\in\{0,1\}^{16},
\]
given by the binary representation of its token ID. We then apply a deterministic lift
\[
\phi:\mathbb{R}^{16}\rightarrow\mathbb{R}^{d_{\text{model}}}
\]
by coordinate repetition, producing the model input
\[
e(t)=\phi(b(t))\in\mathbb{R}^{d_{\text{model}}}.
\]

This lift adds no information. The resulting lookup table in $\mathbb{R}^{d_{\text{model}}}$ has matrix rank at most 16, so the setup is not a learned 1024-dimensional embedding in disguise. Any richer token geometry must therefore be constructed by the trainable Transformer blocks above this fixed interface.

The particular fixed arrangement of repeated coordinates (e.g., contiguous repetition or interleaving) does not change the intrinsic information content or rank of the representation. In the experiments reported here we use a simple deterministic repetition pattern for implementation convenience, and we keep the canonical bit ordering of token IDs throughout.

\subsection{Constructive depth growth}

We train models by constructive growth:

\begin{enumerate}
    \item Initialize a shallow model above the fixed token interface.
    \item Train the current top stack while keeping the token interface fixed.
    \item Freeze the trained dense blocks.
    \item Add a new block group on top.
    \item Train only the newest block group and the LM head, keeping the previously trained dense stack frozen.
\end{enumerate}

The growth chunk size is chosen mechanically to match the active trainable budget rather than tuned for best perplexity. In the controlled 9-layer study we grow in 3-block increments; in the 16-layer long-horizon run we grow in 4-block increments. These choices were budget-driven, not HPO-driven.

\subsection{Budget-preserving global readjustment with LoRA}

For longer runs, freezing the entire lower dense stack can become too restrictive. To allow limited global readjustment without exceeding the same active trainable budget, we interleave LoRA stages between growth phases. During these stages, the dense weights remain frozen and trainable parameters are reallocated to LoRA adapters across all layers.

In the 16-layer long-horizon run, LoRA targets the modules
\texttt{q\_proj}, \texttt{k\_proj}, \texttt{v\_proj}, \texttt{o\_proj}, \texttt{mlp.net.0}, and \texttt{mlp.net.2},
with \texttt{lora\_dropout}=0.05, \texttt{alpha}=r, and \texttt{bias="none"}. The rank $r$ is chosen mechanically to saturate the same active trainable budget, giving $r=797$ at the 1--8 LoRA stage, $r=531$ at the 1--12 LoRA stage, and $r=398$ at the 1--16 LoRA stage. No LoRA is used in the controlled 9-layer comparison, where we intentionally study the stricter frozen-dense setting in isolation.

\subsection{Scope of the claim}

The paper does not claim that this constrained regime matches tuned dense monolithic pretraining, nor that random-initialized stacking is the best generic growth strategy. The object of study is narrower: whether useful continued learning remains possible when the token interface is fixed, previously trained dense blocks remain frozen, and the active trainable set is explicitly bounded throughout training. All empirical conclusions in the paper should be read under this narrower scope.

\FloatBarrier

\section{Experimental Setup}

\subsection{Overview}

We report two main empirical views of the proposed regime. The first is a controlled common-protocol comparison at modest scale, designed to quantify the quality--memory tradeoff against monolithic baselines under matched data, tokenizer, architecture, and training recipe. The second is a longer-horizon experiment above the 16-dim binary token-ID interface, designed as a viability demonstration of continued learning under a bounded active trainable budget.

\subsection{Controlled 9-layer common-protocol study}

Our cleanest comparison uses 9-layer decoder-only Transformers with $d_{\text{model}}=1024$, $n_{\text{head}}=32$, and vocabulary size 65,536. All models are trained on the same 4B-token mixture with the same tokenizer and shared training protocol. The compared variants are:

\begin{itemize}
    \item \textbf{Constructive Frozen (Unicode):} constructive growth above the frozen Unicode interface;
    \item \textbf{Monolithic Frozen (Unicode):} end-to-end monolithic training above the same frozen Unicode interface;
    \item \textbf{Constructive Trainable:} constructive growth with a learned embedding trained only in the first stage and then frozen;
    \item \textbf{Monolithic Trainable:} standard end-to-end monolithic training with a learned embedding;
    \item \textbf{Constructive 16-dim:} constructive growth above the frozen 16-dim binary token-ID interface;
    \item \textbf{Monolithic 16-dim:} end-to-end monolithic training above the same frozen 16-dim binary token-ID interface.
\end{itemize}

We emphasize that this study is a \emph{common-protocol} comparison, not a best-vs-best comparison. We do not perform baseline-specific hyperparameter tuning, so differences should be interpreted as regime differences under a shared protocol rather than as each method at its individually optimized best.

\subsection{Long-horizon 16-layer viability run}

To test whether continued learning remains possible above an extremely minimal fixed interface beyond the 4B-token controlled study, we run a 16-layer decoder-only Transformer with $d_{\text{model}}=1024$, $n_{\text{head}}=32$, and a frozen 16-dim binary token-ID interface on a FineWeb-Edu + Cosmopedia mixture for 68.9B tokens. Training proceeds in 4-layer increments (4$\rightarrow$8$\rightarrow$12$\rightarrow$16), keeping the active trainable set fixed at approximately 117.5M parameters (newest 4 blocks + LM head during growth, or LoRA adapters during global readjustment stages).

Because the data mixture changes across stages through stage-wise instruction/SFT mixing, this experiment should be interpreted as a \emph{long-horizon viability demonstration}, not as a clean causal estimate of the effect of growth alone.

\subsection{Evaluation and interpretation}

We report training loss, perplexity, and selected downstream evaluations including MMLU, ARC-style metrics, and SQuAD exact match. These evaluations are used to test whether useful performance survives under the constraint. They are not intended as a comprehensive capability evaluation, and the paper does not claim statistical superiority or broad competitiveness over tuned dense monolithic baselines.

All controlled comparisons reported here are single-run experiments. Accordingly, we treat small metric differences conservatively and focus on the qualitative question of whether useful continued learning remains possible under the bounded active trainable-parameter regime.

\section{Experiments and Results}

\subsection{Controlled 9-layer common-protocol comparison}

\begin{figure}[htbp]
\begin{center}
\includegraphics[width=0.70\linewidth]{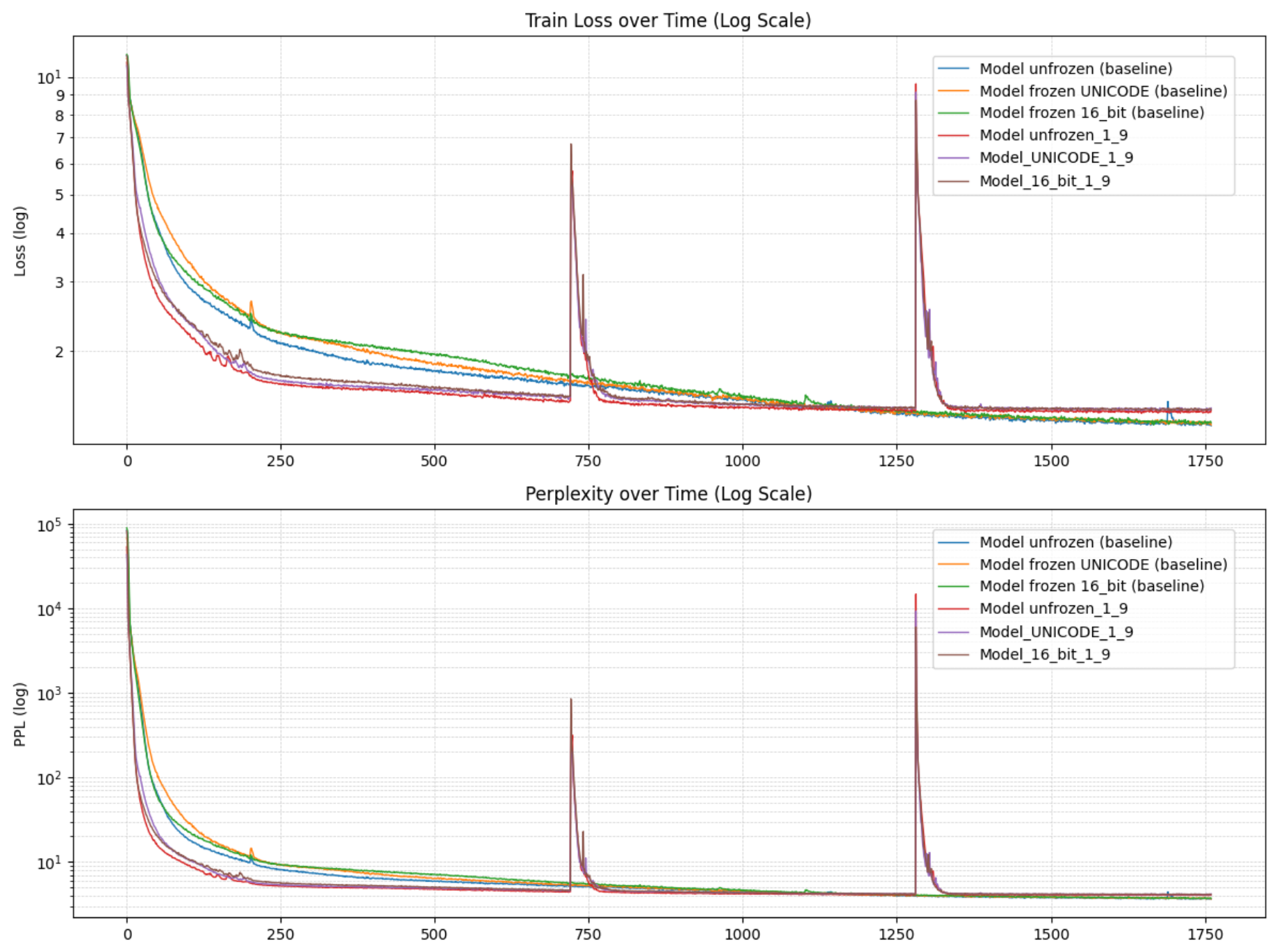}
\end{center}
    \caption{Training dynamics during progressive layer-wise growth for 'Model unfrozen (baseline)', 'Model frozen UNICODE (baseline)', 'Model frozen 16\_dim binary (baseline)', 'Model UNICODE 1\_9', 'Model unfrozen 1\_9' and 'Model 16\_dim binary 1\_9'. Each loss spike marks the stacking of a new layer group, followed by rapid convergence.}
    \label{fig:growth_dynamics_1_9}
\end{figure}

The cleanest comparison in the paper is the 9-layer study, where all models share the same data mixture, tokenizer, architecture, and training recipe. Figure~\ref{fig:growth_dynamics_1_9} shows that constructive growth is stable in this regime: adding a new 3-block group produces a transient loss spike followed by rapid recovery.

Table~\ref{tab:controlled_summary} makes the main tradeoff explicit. For the interface-matched Unicode setting, constructive growth uses 105.0M active trainable parameters, compared with 180.5M for the monolithic frozen Unicode baseline; its final training perplexity is 4.12 versus 3.71 for that baseline. Relative to the fully trainable monolithic baseline, the active trainable set drops from 247.6M to 105.0M. The main result of the controlled study is therefore a quality--memory tradeoff, not parity with monolithic training.

\begin{figure}[htbp]
\begin{center}
\includegraphics[width=0.70\linewidth]{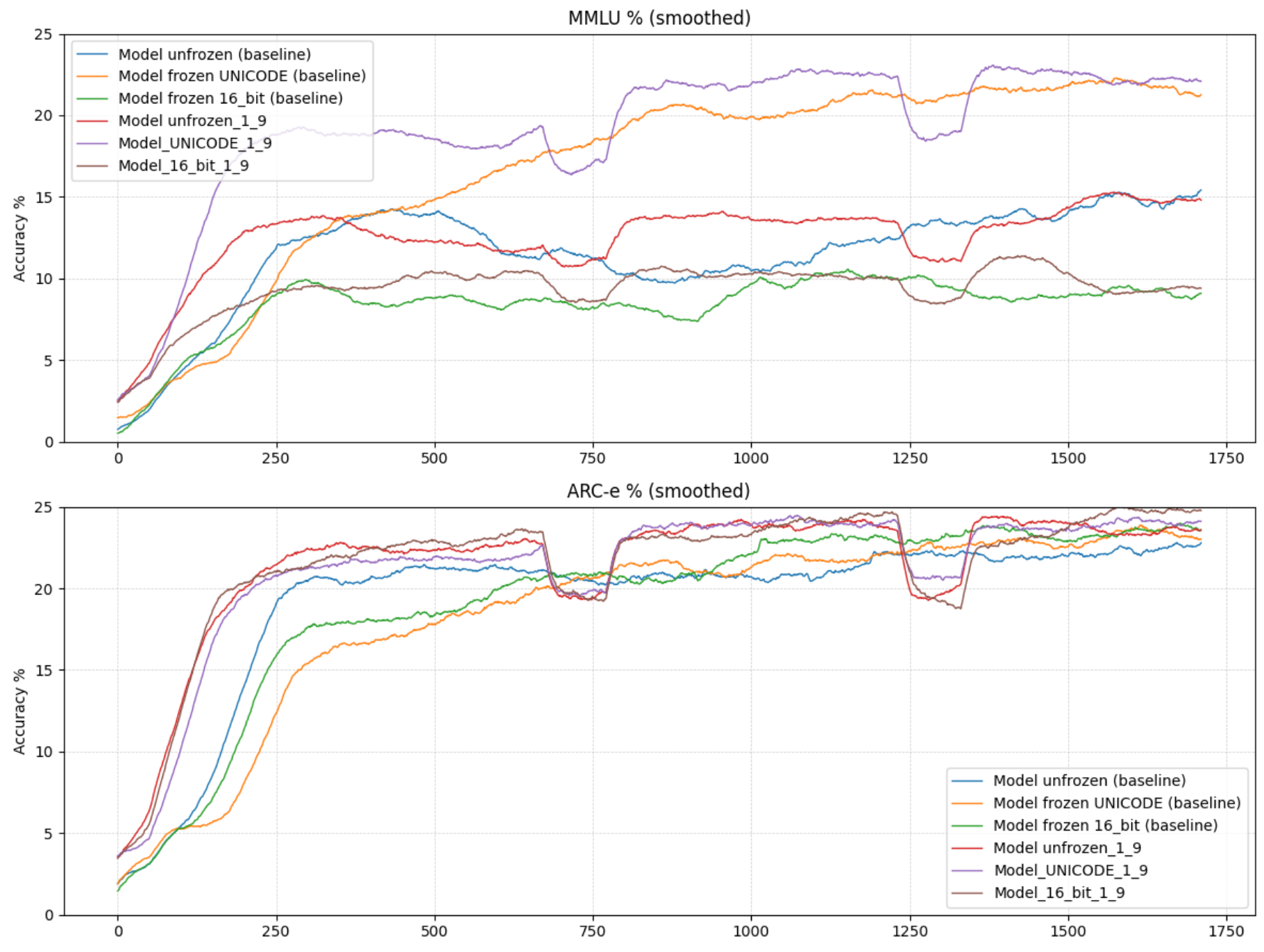}
\end{center}
    \caption{MMLU and ARC-E metric dynamics during progressive layer-wise growth for 'Model unfrozen (baseline)', 'Model frozen UNICODE (baseline)', 'Model frozen 16\_dim binary (baseline)', 'Model UNICODE 1\_9', 'Model unfrozen 1\_9' and 'Model 16\_dim binary 1\_9'.}
    \label{fig:mmlu_arc_e_dynamics_1_9}
\end{figure}

Selected downstream metrics are shown in Figure~\ref{fig:mmlu_arc_e_dynamics_1_9}. The constructive Unicode model remains in the same broad performance range as the monolithic baselines on the reported evaluations, but because all results are single-run, we do not interpret small differences as statistically established. The constructive trainable-embedding ablation does not outperform its monolithic counterpart in this regime. This suggests that a fixed bottom interface may be compatible with constructive growth, but it does not establish a formal interaction effect.

Constructive growth also remains viable above the 16-dim binary token-ID interface. Despite the extreme rank constraint of this fixed input representation, the constructive 16-dim binary model converges stably and retains non-trivial downstream performance. We view this as evidence that useful continued learning can proceed above a minimal fixed interface.

\begin{figure}[htbp]
\begin{center}
\includegraphics[width=0.70\linewidth]{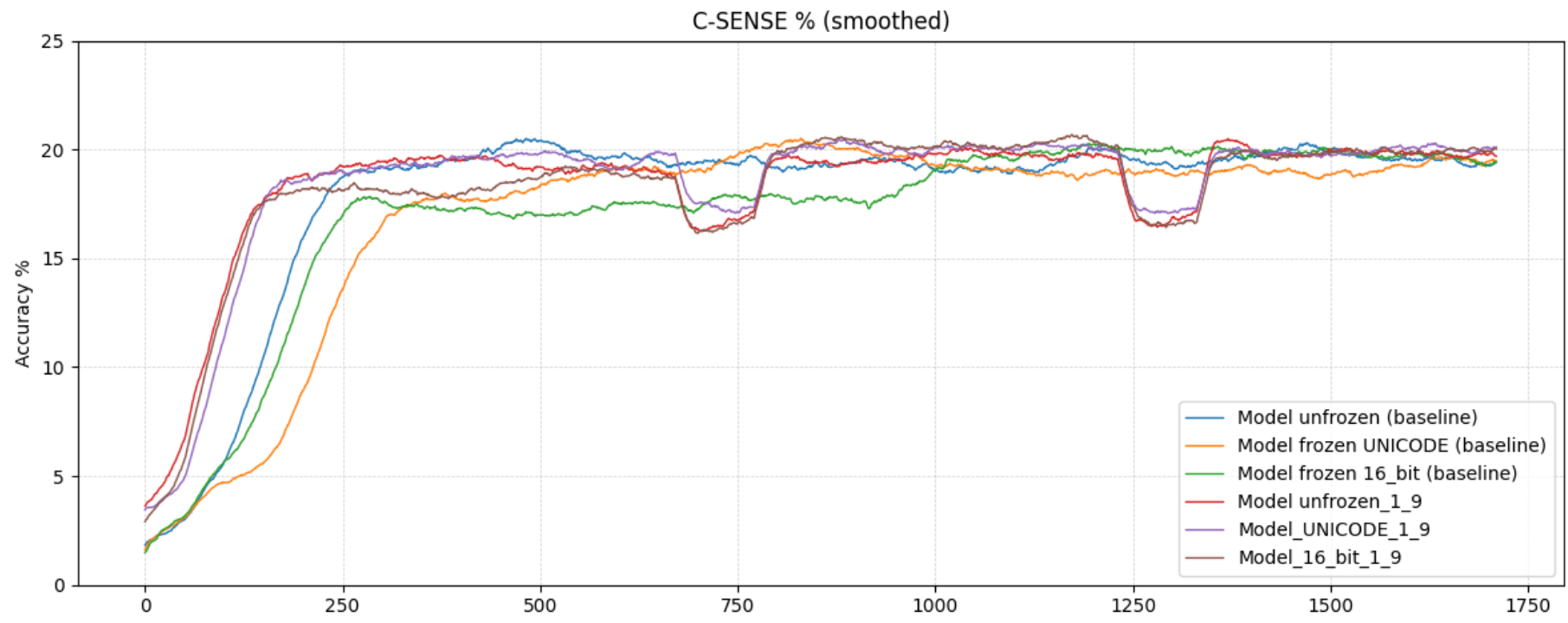}
\end{center}
    \caption{C-SENSE (CommonsenseQA accuracy) metric dynamics during progressive layer-wise growth for 'Model unfrozen (baseline)', 'Model frozen UNICODE (baseline)', 'Model frozen 16\_dim binary (baseline)', 'Model UNICODE 1\_9', 'Model unfrozen 1\_9' and 'Model 16\_dim binary 1\_9'.}
    \label{fig:c_sense_arc_c_dynamics_1_9}
\end{figure}

\begin{table}[t]
\centering
\caption{Controlled 9-layer common-protocol comparison at the final stage.}
\label{tab:controlled_summary}
\small
\begin{tabular}{lccc}
\toprule
Model & Active trainable (M) & Frozen (M) & Final training PPL \\
\midrule
Monolithic Trainable & 247.6 & 0.0 & 3.77 \\
Monolithic Frozen (Unicode) & 180.5 & 67.1 & 3.71 \\
Constructive Frozen (Unicode) & 105.0 & 142.7 & 4.12 \\
Constructive Trainable & 105.0 & 142.7 & 4.04 \\
Monolithic Frozen (16-dim binary) & 180.5 & 1.0 & 3.72 \\
Constructive (16-dim binary) & 105.0 & 76.6 & 4.16 \\
\bottomrule
\end{tabular}
\end{table}

\FloatBarrier

\subsection{Long-horizon 16-layer viability run}

Appendix~\ref{app:edu16_scaling} reports the full schedule and hyperparameters. In the final merged 269.7M base model, the 1--16 LoRA stage reaches 28.92\% MMLU after 68.9B training tokens. During LoRA training, temporary adapter parameters raise the instantaneous parameter count, but after merging the adapters the inference-time model returns to the 269.7M base size. Because the stage-wise SFT mixture changes across phases, we interpret this experiment as viability evidence rather than a clean causal comparison.

\section{Discussion}

\subsection{What the paper does and does not show}

The paper should be read as a study of a constrained training regime, not as evidence that constructive growth is generally better than dense monolithic pretraining. In the controlled 9-layer comparison, monolithic training remains stronger on final perplexity. The result we emphasize is narrower: substantial reductions in active trainable parameters can be achieved while preserving useful continued learning above a fixed token interface.

This distinction matters for practitioners. If a training setup can comfortably support tuned dense monolithic pretraining with optimizer-state sharding and full-model updates, that remains the natural quality-oriented baseline. Our method is instead relevant to settings where one explicitly wants to bound the active trainable set and avoid reopening the full dense stack during continued training.

\subsection{The role of a fixed minimal interface}

The most distinctive aspect of this paper is not growth by itself, but growth above a fixed and sometimes extremely minimal interface. The 16-dim binary token-ID setting is an intentionally severe stress test: before the first trainable block, the model only receives a deterministic lift of a 16-dimensional binary code. In that sense, the setup removes most of the usual freedom associated with learned high-capacity embeddings.

The fact that useful continued learning remains possible in this setting suggests that semantically rich trainable input embeddings are not a strict prerequisite for forming useful internal representations. This does not imply that learned embeddings are unnecessary in general, nor that the 16-dim binary interface is an optimal practical design. Rather, it shows that meaningful token geometry can be induced by the trainable Transformer stack above an extremely limited fixed interface.

\subsection{Freezing lower dense blocks and partial global readjustment}

A clear limitation of strict constructive growth is that early dense representations cannot be directly revised once frozen. In longer schedules this can make the model increasingly rigid. Our use of interleaved LoRA stages is intended as a partial remedy: it provides a limited path for global readjustment while preserving the same active trainable-parameter budget. However, this is still not equivalent to reopening the dense model. The paper therefore studies a deliberately stronger constraint than standard continued pretraining.

\subsection{Relation to prior growth methods}

Existing work on progressive stacking, copy-based growth, and function-preserving expansion often focuses on improving optimization or preserving function when moving to a larger model. That literature is important and should be treated as the primary backdrop for this paper. Our narrower question is different: how much continued learning remains possible when the token interface is fixed, previously trained dense blocks remain frozen, and the active trainable set is explicitly bounded throughout training? Under this reading, the paper is best viewed as an empirical study of a constrained regime rather than as a new generic growth algorithm.

\subsection{Future work}

Several follow-up directions would materially strengthen this line of work. The most important are: multi-seed controlled studies; baseline-specific tuning for monolithic baselines; direct comparisons against copy-based growth methods such as G\_stack / Net2Net-style expansion; matched with/without-LoRA comparisons under identical data schedules; chunk-size sweeps; alternative layer-selection policies under the same active budget; and controlled studies of code assignments or token-ID permutations in the 16-dim binary setting. A broader capability evaluation beyond the selected benchmarks used here would also be valuable.

\section{Limitations}

This paper should be read as a feasibility/tradeoff study rather than a best-vs-best comparison against tuned dense monolithic pretraining.

First, the controlled comparisons are based on single runs, so small metric differences should not be interpreted as statistically established. Second, the 9-layer comparison uses a common training protocol rather than baseline-specific hyperparameter tuning; it is therefore a regime comparison, not an individually optimized comparison. Third, the 4B-token controlled study uses a mixed corpus rather than a standard large-scale pure pretraining corpus, which may affect absolute benchmark levels.

Fourth, the 68.9B-token long-horizon run changes the data mixture across stages through increasing instruction/SFT mixing, which confounds causal attribution and makes that experiment a viability demonstration only. Fifth, we do not compare against several important growth baselines, including copy-based or function-preserving expansion strategies (e.g., G\_stack / Net2Net-style growth), alternative layer-selection policies, or matched with/without-LoRA schedules under identical conditions.

Sixth, the LoRA stages were validated in a standard training regime with FP32
optimizer states. The same ranks, stability, or quality tradeoffs may not
transfer directly to other systems regimes such as 8-bit optimizers,
optimizer-state sharding, or different mixed-precision settings. In addition,
the LoRA ranks in our long-horizon runs were chosen mechanically to saturate
the active trainable budget rather than tuned as standard PEFT hyperparameters.

Seventh, the 16-dim binary token-ID experiments use one deterministic code assignment, namely the canonical binary encoding of token IDs. A systematic comparison of alternative code assignments or token-ID permutations is left for future work. Finally, freezing the lower dense stack limits the ability to revise early representations. Interleaved LoRA provides only a partial global readjustment path and is not equivalent to reopening the full dense model. We also evaluate only selected benchmarks rather than broader generation quality or capability coverage.

Because the tokenizer uses a structured ID assignment (Unicode ranges, reserved regions, and dedicated n-gram regions), the canonical 16-dim binary token-ID code is not a random hash over the vocabulary. It may therefore retain weak surface regularities inherited from the tokenizer design. The present paper does not isolate this effect with systematic token-ID permutation or random reassignment experiments.

Any injective deterministic lift of the 16-bit code to $\mathbb{R}^{d_{\text{model}}}$ preserves the same intrinsic 16-bit information content. For simple repetition patterns such as contiguous repetition versus interleaving, the resulting $d_{\text{model}}$-dimensional representations differ only by a fixed coordinate permutation. We therefore do not expect a change in intrinsic information content or representational rank, although optimization dynamics may still differ; this effect is not studied systematically here.

\section{Conclusion}

We presented a constrained training regime for decoder-only Transformers in which a fixed deterministic token interface is held constant, previously trained dense blocks are not reopened, and the active trainable-parameter set remains bounded as depth grows. The main result is not that this regime matches dense monolithic pretraining---it does not on perplexity in our controlled comparison---but that continued learning remains possible under this constraint, even above a minimal 16-dim binary token-ID code.

We view this as evidence for a narrow but useful claim: in the regimes studied
here, continued depth growth can proceed without a trainable input embedding
or dense full-model reopening. 

\section{Impact Statement}

This paper studies a training methodology for decoder-only Transformers under explicit memory and optimization constraints. Potential benefits include reducing peak optimizer-state memory, making some forms of continued training more accessible to smaller labs, and enabling controlled scientific studies of how useful internal representations can arise above fixed token interfaces.

\bibliographystyle{plainnat}
\bibliography{main}



\newpage
\appendix
\onecolumn

\FloatBarrier
\section{APPENDIX: Training loss and perplexity data}

\begin{table}[ht!] 
\centering 
\caption{Training loss and perplexity data}
\label{tab:loss_table}
\small 
\begin{tabular}{rrrrrrrrrrrrr}
\toprule
Step& 
{\scriptsize \shortstack{Model\\unfrozen\\(base\\line)\\Loss}} & 
{\scriptsize \shortstack{Model\\unfrozen\\(base\\line)\\PPL}} & 
{\scriptsize \shortstack{Model\\frozen\\UNI\\CODE\\(base\\line)\\Loss}} & 
{\scriptsize \shortstack{Model\\frozen\\UNI\\CODE\\(base\\line)\\PPL}} & 
{\scriptsize \shortstack{Model\\frozen\\16\_dim\\binary\\(base\\line)\\Loss}} & 
{\scriptsize \shortstack{Model\\frozen\\16\_dim\\binary\\(base\\line)\\PPL}} &
{\scriptsize \shortstack{Model\\unfrozen\\1\_9\\Loss}} & 
{\scriptsize \shortstack{Model\\unfrozen\\1\_9\\PPL}} & 
{\scriptsize \shortstack{Model\\UNI\\CODE\\1\_9\\Loss}} & 
{\scriptsize \shortstack{Model\\UNI\\CODE\\1\_9\\PPL}} & 
{\scriptsize \shortstack{Model\\16\_dim \\binary\\1\_9\\Loss}} & 
{\scriptsize \shortstack{Model\\16\_dim \\binary\\1\_9\\PPL}} \\

\midrule
0 & 11.34 & 84005 & 11.12 & 67363 & 11.40 & 89155 & 10.88 & 53254 & 10.67 & 43094 & 11.32 & 82104 \\
050000 & 2.90 & 18.27 & 3.35 & 28.62 & 3.11 & 22.50 & 2.21 & 9.11 & 2.36 & 10.63 & 2.37 & 10.73 \\
100000 & 2.29 & 9.82 & 2.47 & 11.85 & 2.38 & 10.85 & 1.72 & 5.61 & 1.76 & 5.79 & 1.84 & 6.29 \\
150000 & 2.00 & 7.38 & 2.14 & 8.48 & 2.15 & 8.60 & 1.63 & 5.08 & 1.65 & 5.22 & 1.69 & 5.43 \\
200000 & 1.86 & 6.45 & 2.00 & 7.35 & 2.06 & 7.85 & 1.59 & 4.93 & 1.61 & 5.03 & 1.66 & 5.24 \\
250000 & 1.77 & 5.90 & 1.85 & 6.37 & 1.96 & 7.10 & 1.56 & 4.78 & 1.60 & 4.94 & 1.60 & 4.95 \\
300000 & 1.71 & 5.52 & 1.75 & 5.76 & 1.84 & 6.28 & 1.53 & 4.62 & 1.55 & 4.73 & 1.57 & 4.83 \\
350000 & 1.66 & 5.24 & 1.70 & 5.45 & 1.74 & 5.71 & 1.50 & 4.48 & 1.53 & 4.63 & 1.54 & 4.67 \\
400000 & 1.60 & 4.98 & 1.63 & 5.11 & 1.66 & 5.26 & 1.47 & 4.37 & 1.51 & 4.51 & 1.52 & 4.58 \\
450000 & 1.54 & 4.69 & 1.56 & 4.77 & 1.58 & 4.86 & 1.45 & 4.27 & 1.49 & 4.41 & 1.49 & 4.44 \\
500000 & 1.50 & 4.48 & 1.50 & 4.49 & 1.53 & 4.63 & 1.44 & 4.22 & 1.47 & 4.33 & 1.46 & 4.32 \\
550000 & 1.46 & 4.29 & 1.47 & 4.34 & 1.49 & 4.42 & 1.42 & 4.12 & 1.44 & 4.22 & 1.44 & 4.22 \\
600000 & 1.40 & 4.06 & 1.40 & 4.04 & 1.43 & 4.17 & 1.42 & 4.12 & 1.44 & 4.21 & 1.43 & 4.19 \\
650000 & 1.38 & 3.96 & 1.39 & 4.02 & 1.39 & 4.03 & 2.25 & 9.45 & 2.02 & 7.53 & 2.02 & 7.54 \\
700000 & 1.35 & 3.87 & 1.37 & 3.92 & 1.37 & 3.93 & 1.41 & 4.09 & 1.43 & 4.18 & 1.42 & 4.14 \\
750000 & 1.33 & 3.78 & 1.34 & 3.80 & 1.35 & 3.85 & 1.41 & 4.09 & 1.43 & 4.17 & 1.42 & 4.15 \\
800000 & 1.31 & 3.71 & 1.32 & 3.75 & 1.35 & 3.86 & 1.40 & 4.06 & 1.42 & 4.14 & 1.42 & 4.16 \\
850000 & 1.33 & 3.77 & 1.31 & 3.71 & 1.31 & 3.72 & 1.40 & 4.04 & 1.42 & 4.12 & 1.43 & 4.16 \\
\bottomrule

\end{tabular}
\end{table}

\FloatBarrier
\section{APPENDIX: Parameter Details for Controlled Study Models}
\label{app:small_model_params}

Table \ref{tab:small_model_growth} provides the corresponding parameter breakdown for the smaller-scale models used in our controlled comparative study. The base architecture for these models is $d_{\text{model}}=1024$, $n_{\text{head}}=32$, and a vocabulary size of 65,536. The models were grown in stages of three layers each, with previously trained layers being frozen at subsequent stages.

\begin{table}[ht!]
\centering
\caption{Parameter breakdown for the controlled-study models. Final sizes are 247.6M for the Unicode/trainable-interface models and 181.6M for the 16-dim binary-interface models.}
\label{tab:small_model_growth}
\begin{tabular}{@{}llccc@{}}
\toprule
\textbf{Model Type} & \textbf{Stage (Layers)} & \textbf{Total (M)} & \textbf{Trainable (M)} & \textbf{Frozen (M)} \\
\midrule
\multirow{3}{*}{\begin{tabular}[c]{@{}l@{}}Constructive (UNICODE) \\ \textit{(Our Method)}\end{tabular}} & 1-3 & 172.1 & 105.0 & 67.1 \\
 & 1-6 & 209.8 & 105.0 & 104.9 \\
 & \textbf{1-9 (Final)} & \textbf{247.6} & \textbf{105.0} & \textbf{142.7} \\
\midrule
\multirow{3}{*}{\begin{tabular}[c]{@{}l@{}}Constructive (Trainable Emb.) \\ \textit{(Constructive trainable)}\end{tabular}} & 1-3 & 172.1 & 172.1 & 0.0 \\
 & 1-6 & 209.8 & 105.0 & 104.9 \\
 & \textbf{1-9 (Final)} & \textbf{247.6} & \textbf{105.0} & \textbf{142.7} \\
\midrule
\multirow{3}{*}{\begin{tabular}[c]{@{}l@{}}Constructive (16-dim binary Emb.) \\ \textit{(Ablation)}\end{tabular}} & 1-3 & 106.0 & 105.0 & 1.0 \\
 & 1-6 & 143.8 & 105.0  & 38.8\\
 & \textbf{1-9 (Final)} & \textbf{181.6} & \textbf{105.0} & \textbf{76.6} \\
 
\midrule
\begin{tabular}[c]{@{}l@{}}Monolithic (Trainable Emb.) \\ \textit{(Baseline trainable classic)}\end{tabular} & \textbf{1-9 (Final)} & \textbf{247.6} & \textbf{247.6} & \textbf{0.0} \\

\midrule
\begin{tabular}[c]{@{}l@{}}Monolithic (Frozen UNICODE Emb.) \\ \textit{(Baseline)}\end{tabular} & \textbf{1-9 (Final)} & \textbf{247.6} & \textbf{180.5} & \textbf{67.1} \\

\midrule
\begin{tabular}[c]{@{}l@{}}Monolithic (Frozen 16\_dim binary Emb.) \\ \textit{(Baseline)}\end{tabular} & \textbf{1-9 (Final)} & \textbf{181.6} & \textbf{180.5} & \textbf{1.0} \\

\bottomrule
\end{tabular}
\end{table}

\FloatBarrier
\section{APPENDIX: Extended 16-dim binary (n\_embed = 16) scaling run on FineWeb-Edu + Cosmopedia}
\label{app:edu16_scaling}
\paragraph{Setup.}
We train a decoder-only Transformer with $d_{\text{model}}=1024$, $n_{\text{head}}=32$, dropout=0.05, and a frozen 16-dimensional binary token-ID embedding (expanded to $d_{\text{model}}$ by repetition). Training proceeds via constructive growth in 4-layer increments. At each growth stage, previously trained blocks remain frozen; only the newest 4 blocks and the LM head are trained, keeping trainable parameters fixed at $\approx$117.5M. Between growth stages we optionally run a LoRA phase applied to all blocks, choosing ranks such that the total number of trainable LoRA parameters also stays within $\approx$117.5M. The training mixture is based on FineWeb-Edu and Cosmopedia (web-scale), with an interleaved instruction-tuning (SFT) dataset sampled either periodically or with probability $p_{\text{sft}}$ (see Table~\ref{tab:edu16_hparams}).
\begin{table}[H]
\centering
\caption{Extended 16-dim binary  (n\_embed = 16) scaling run: stage schedule, parameter counts, and training budget. Trainable parameters are kept approximately constant at $\approx$117.5M across stages (either newest 4 blocks + LM head, or LoRA adapters). Tokens seen are taken from the training log (end of stage). For LoRA stages, the ``Total'' column includes temporary adapter parameters during training. After merging the adapters into the frozen base weights for evaluation, the inference-time model returns to the corresponding base-model size (e.g., 269.7M at the final 16-layer stage).
}
\label{tab:edu16_schedule}
\small
\begin{tabular}{lcccccc}
\toprule
Stage & Layers & Method & Total (M) & Trainable (M) & Frozen (M) & Tokens seen (B) \\
\midrule
1--4   & 4  & Growth & 118.6 & 117.5 & 1.0   & 15.5 \\
1--8   & 8  & Growth & 169.0 & 117.5 & 51.4  & 31.0 \\
1--8   & 8  & LoRA   & 286.5 & 117.5 & 169.0 & 38.8 \\
1--12  & 12 & Growth & 219.3 & 117.5 & 101.8 & 54.3 \\
1--12  & 12 & LoRA   & 336.8 & 117.4 & 219.3 & 59.2 \\
1--16  & 16 & Growth & 269.7 & 117.5 & 152.2 & 65.5 \\
1--16  & 16 & LoRA   & 387.1 & 117.4 & 269.7 & 68.9 \\
\bottomrule
\end{tabular}
\end{table}

\begin{table}[H]
\centering
\caption{Extended 16-dim binary scaling run: downstream metrics by stage (reported as mean as logged during training).}
\label{tab:edu16_metrics}
\small
\begin{tabular}{lcc}
\toprule
Stage & MMLU & SQuAD \\
\midrule
1--4 Growth   & 16.95 $\pm$ 0.16 & 1.21 $\pm$ 0.40 \\
1--8 Growth   & 19.42 $\pm$ 0.16 & 2.77 $\pm$ 0.48 \\
1--8 LoRA     & 21.31 $\pm$ 0.14 & 5.39 $\pm$ 0.62 \\
1--12 Growth  & 23.38 $\pm$ 0.16 & 9.45 $\pm$ 0.83 \\
1--12 LoRA    & 23.90 $\pm$ 0.19 & 12.19 $\pm$ 0.80 \\
1--16 Growth  & 24.58 $\pm$ 0.18 & 13.48 $\pm$ 0.94 \\
1--16 LoRA    & 28.92 $\pm$ 0.19 & 19.10 $\pm$ 0.95 \\
\bottomrule
\end{tabular}
\end{table}

\begin{table}[H]
\centering
\caption{Key hyperparameters for the extended 16-dim binary scaling run (stage-wise). We keep the effective token batch approximately constant by adjusting gradient accumulation when micro-batch size changes.}
\label{tab:edu16_hparams}
\small
\begin{tabular}{lcccc}
\toprule
Stage & Micro-batch & Grad accum & LR & SFT mixing \\
\midrule
1--4 Growth   & 108 & 32 & $1.05 \cdot 8\mathrm{e}{-4}$ & periodic (every 250 iters) \\
1--8 Growth   & 108 & 32 & $0.95 \cdot 8\mathrm{e}{-4}$ & periodic (every 250 iters) \\
1--8 LoRA     & 64  & 54 & $0.50 \cdot 8\mathrm{e}{-4}$ & periodic (every 250 iters) \\
1--12 Growth  & 108 & 32 & $0.45 \cdot 8\mathrm{e}{-4}$ & $p_{\text{sft}}=0.004$ \\
1--12 LoRA    & 48  & 72 & $0.25 \cdot 8\mathrm{e}{-4}$ & $p_{\text{sft}}=0.01$ \\
1--16 Growth  & 104 & 32 & $0.10 \cdot 8\mathrm{e}{-4}$ & $p_{\text{sft}}=0.05$ \\
1--16 LoRA    & 36  & 96 & $0.055 \cdot 8\mathrm{e}{-4}$ & $p_{\text{sft}}=0.33$ \\
\bottomrule
\end{tabular}
\end{table}


\end{document}